\documentclass[runningheads]{llncs}
\usepackage{graphicx}

\usepackage{tikz}
\usepackage{comment}
\usepackage{amsmath,amssymb} 
\usepackage{color}

\begin{document}

\title{Edge-Preserving Guided Semantic Segmentation for VIPriors Challenge} 

\author{Chih-Chung Hsu \and
Hsin-Ti Ma}

\institute{National Pingtung University of Science and Technology, \\
No.1, Shufu Rd., Neipu, Pingtung country, Taiwan. \\
\email{cchsu@mail.npust.edu.tw}\\
\url{https://cchsu.info}}

\maketitle

\begin{abstract}
Semantic segmentation is one of the most attractive research fields in computer vision. In the VIPriors challenge, only very limited numbers of training samples are allowed, leading to that the current state-of-the-art and deep learning-based semantic segmentation techniques are hard to train well. To overcome this shortcoming, therefore, we propose edge-preserving guidance to obtain the extra prior information, to avoid the overfitting under small-scale training dataset. First, a two-channeled convolutional layer is concatenated to the last layer of the conventional semantic segmentation network. Then, an edge map is calculated from the ground truth by Sobel operation and followed by concatenating a hard-thresholding operation to indicate whether the pixel is the edge or not. Then, the two-dimensional cross-entropy loss is adopted to calculate the loss between the predicted edge map and its ground truth, termed as an edge-preserving loss. In this way, the continuity of boundaries between different instances can be forced by the proposed edge-preserving loss. Experiments demonstrate that the proposed method can achieve excellent performance under small-scale training set, compared to state-of-the-art semantic segmentation techniques.

\end{abstract}

\section{Introduction}
With the rapid growth of deep learning techniques, several semantic segmentation models were proposed recently. The strongest baseline for semantic segmentation, as well-known, is the DeepLabV3+ \cite{deeplab}. The encoder-decoder architecture makes the semantic information can be preserved from middle- and high-level feature maps, leading to state-of-the-art performance on various benchmarks. Since the dataset of the VIPriors challenges is a subset of cityscape \cite{cityscape}, several state-of-the-art semantic segmentation techniques are selected as the candidates, such as DeepLabV3+ \cite{deeplab}, Densely Connected Neural Architecture Search (DCNAS) \cite{dcans}, Object-Contextual Representation (OCR) \cite{hrnet}, Height-Driven Attention Networks (HANet) \cite{hanet}. Since the network architecture search (NAS) is widely used to find the optimized network architecture for various tasks. However, such a NAS-based network lacks generalization so that the performance of the different datasets may not be promising. OCR discovers the context information from objects/instances to obtain the promising feature representation to further boost the performance for semantic segmentation. Both DeepLabV3+ and OCR achieve state-of-the-art performance. However, the number of the parameters of these two models is relatively large, leading to the fact that the performance under small-scale training datasets maybe not promising. 

Recently, HANet \cite{hanet} introduces the extra image prior to the learning process--height-driven prior--to achieve a significant performance gain for existing semantic segmentation techniques. However, the number of parameters of HANet is still large, so that it may not be suitable for a small-scale dataset. Inspired by HANet \cite{hanet}, we propose the edge-preserving loss to introduce more images prior to making the network can be well-trained on the limited dataset. Based on HANet, we make three improvements to boost the performance under a small-scale dataset. Our contribution is three-fold:
\begin{itemize}
    \item We introduce the edge-preserving loss to force the contours of instances to ensure the continuity of the predicted segmentation maps.
    \item To have better feature representation powerful, we introduce the ResNeSt \cite{resnest} to the HANet as the backbone without dramatically increasing the number of parameters.
    \item A simple but effective data augmentation strategy is proposed to solve the label unbalanced issue. 
\end{itemize}

	The rest of this paper is organized as follows. Section 2 introduces the proposed semantic segmentation method. Section 3 presents obtained experimental results of the mini-Cityscape. Finally, Section 4 gives the~conclusions.

\section{Method}
	\begin{figure}
		\centering
		\includegraphics[width=1\textwidth]{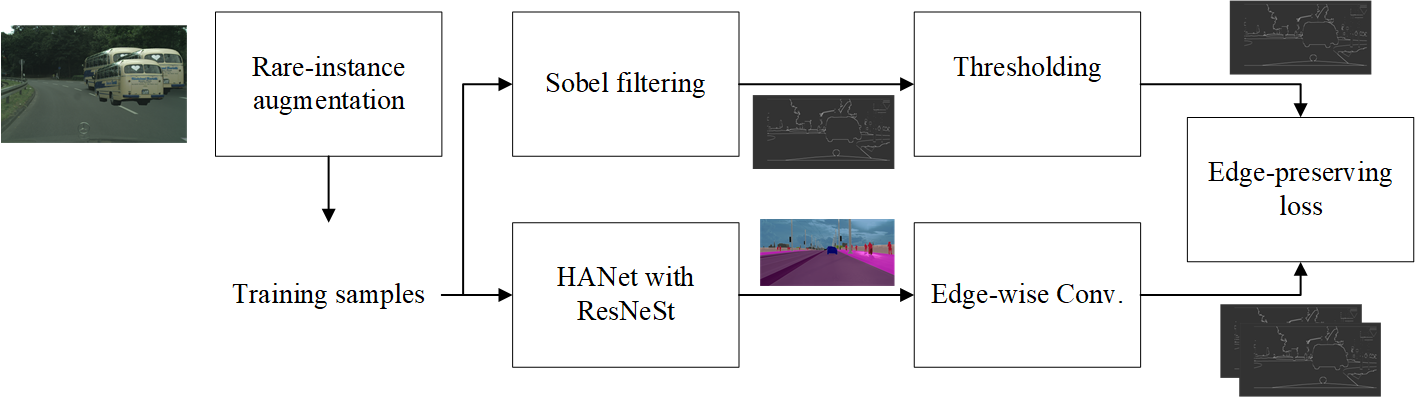}
		\caption{The flowchart of the proposed edge-preserving semantic segmentation method.}
		\label{fig:flowchart}	
	\end{figure}
The flowchart of the proposed method is depicted in Fig. \ref{fig:flowchart}. First, the probability of each class in the mini-Cityscape is calculated. We observed that some of the classes are relatively rare, such as the wall, fence, bus, and train. With these classes, we propose a simple data augmentation technique to increase the probability of those classes. Next, we adopt the ResNeSt as the backbone network with dilated convolution with the second split-attention block in ResNeSt. Third, the edge map of the predicted segmentation map is calculated by concatenating an extra convolutional layer with two $3\times 3$ kernels. Meanwhile, the Sobel filtering is applied to the ground truth to obtain the filtered edge map and followed by applying a thresholding function to obtain the binary edge map. By minimizing the cross-entropy loss between the edge maps of the predicted and ground truth, the HANet can be guided without significantly overfitting.

\textbf{Data Augmentation}. In addition to the standard data augmentation operations, we propose a simple but effective data augmentation method for rare instances. First, for the rare instances such as the wall, fence, bus, and train, each instance is randomly resized and flipped horizontally and followed by pasting to the original images. In this way, it is easy to increase the pixels of rare instances. Our experiments show the effectiveness of the proposed data augmentation technique.

\textbf{ResNeSt as Backbone}. The better the backbone network is the more significant improvement of performance we have. Since the best performance of the backbone network is ResNeSt \cite{resnest}, it is a natural choice to adopt ResNeSt as our backbone network on HRNet. To have a bigger receptive field in the middle layers of the backbone, we set the dilation coefficient of the ResNeSt to $2$ to obtain bigger feature maps. 

\textbf{Edge-preserving loss}. Since it is hard to learn the complete and fine-detailed segmentation maps from the limited training set, we should introduce more priors to learning architecture to avoid the overfitting. Therefore, we first adopt the Sobel filter on the ground truth to obtain the edge map. Since the edge map is not label-wise (i.e., floating-point data type), it is hard to optimize the network by such complex information. Instead, a thresholding function is used to make the filtered edge map to be a binary edge map by truncating the values, which is larger than 1 to be 1. Since the hard-thresholding function is non-differentiable, we can not apply the thresholding function on the predicted segmentation map. Instead, we add an extra convolutional layer with two $3\times 3$ kernels to make the predicted segmentation map to be a binary classification map. The standard two-dimensional cross-entropy loss is adopted as the metric for network training. In this way, we can not only force the continuity of the boundaries between instances but also can ensure the compacted prediction results. 

\section{Experiments}
In the experiments, the mini-Cityscape dataset is used. It is a subset of Cityspcpe dataset. The number of the training samples is $200$ and there are $19$ classes. The crop size of the proposed method is $512$ and the batch size is $3$ per GPU. The learning rate is $0.01$ with proposed method is $512$ and the batch size is $3$ per GPU. The learning rate is $0.01$ with standard SGD optimizer. The most parameters of HANet is set to default values. The total epoch of the training process is $100$. We evaluate the performance on the validation set of mini-Cityscape. Table \ref{tab:result1} presents the performance comparison between the proposed method and the baseline HRNet for semantic segmentation. As a result, our method can somewhat improve the performance under the small-scale training set.

    \begin{table}
      \caption{Performance comparison on mini-Cityscape dataset}
      
      \centering
      \begin{tabular}{c|c|c|c|c}
        \hline
        Method & IOU$_{cls}$ & mIoU$_{cls}$ & IoU$_{cat}$& IoU$_{cat}$\\
        \hline
        HANet\cite{hanet}        & 0.513 & 0.277 & 0.617 & 0.522 \\
        HANet+DA                & 0.548 & 0.307 & 0.809 & 0.586 \\
        HANet+DA+ResNeSt       & 0.631 & 0.398 & 0.872 & 0.701 \\
        Proposed method           & \textbf{0.644} & \textbf{0.405} & \textbf{0.873} & \textbf{0.705} \\
        \hline
    \end{tabular}\label{tab:result1}
    \end{table}

\section{Conclusions}
In this paper, a novel edge-preserving loss has proposed to keep the continuity of the boundaries between instances in the predicted segmentation map. Since the number of the training samples is relative low, it is hard to train a generalized network to well predict the segmentation map. With the proposed edge-preserving loss, we can embed the image intrinsic priors to the learning process, leading to a fact that the proposed method is suitable for the small-scale dataset. Furthermore, we also demonstrated that the proposed simple data augmentation strategy by duplicating the rare instances is effective way to boost the performance. Experimental results have verified the effectiveness of the proposed method. 

%
%
\bibliographystyle{splncs04}
\bibliography{egbib}
\end{document}